\documentclass[sigchi,nonacm,screen]{acmart}

\usepackage{array}
\AtBeginDocument{%
  \providecommand\BibTeX{{%
    \normalfont B\kern-0.5em{\scshape i\kern-0.25em b}\kern-0.8em\TeX}}}

\setcopyright{rightsretained}
\copyrightyear{2021}

\newcommand{\systemname}{\textsc{imacs}~}
\newcommand{\systemnamenospace}{\textsc{imacs}}

\usepackage{enumitem}

\newenvironment{itemizecompact}{
    \begin{itemize}
        \setlength{\itemsep}{1pt}
        \setlength{\parskip}{0pt}
        \setlength{\parsep}{0pt}
}{
    \end{itemize}
}

\begin{document}

\title{IMACS: Image Model Attribution Comparison Summaries}

\author{Eldon Schoop}
\email{eschoop@berkeley.edu}
\authornote{Work done while interning at Google.}
\affiliation{%
  \institution{UC Berkeley EECS}
  \city{Berkeley}
  \state{CA}
  \country{USA}
}
\author{Ben Wedin}
\email{wedin@google.com}
\affiliation{%
  \institution{Google Research}
  \city{Cambridge}
  \state{MA}
  \country{USA}
}
\author{Andrei Kapishnikov}
\email{kapishnikov@google.com}
\affiliation{%
  \institution{Google Research}
  \city{Cambridge}
  \state{MA}
  \country{USA}
}
\author{Tolga Bolukbasi}
\email{tolgab@google.com}
\affiliation{%
  \institution{Google Research}
  \city{Cambridge}
  \state{MA}
  \country{USA}
}
\author{Michael Terry}
\email{michaelterry@google.com}
\affiliation{%
  \institution{Google Research}
  \city{Cambridge}
  \state{MA}
  \country{USA}
}

\begin{abstract}
    Developing a suitable Deep Neural Network (DNN) often requires significant iteration, where different model versions are evaluated and compared. While metrics such as accuracy are a powerful means to succinctly describe a model's performance across a dataset or to directly compare model versions, practitioners often wish to gain a deeper understanding of the factors that influence a model's predictions. Interpretability techniques such as gradient-based methods and local approximations can be used to examine small sets of inputs in fine detail, but it can be hard to determine if results from small sets generalize across a dataset. We introduce IMACS, a method that combines gradient-based model attributions with aggregation and visualization techniques to
    summarize differences in attributions between two DNN image models. 
    More specifically, IMACS extracts salient input features from an evaluation dataset, clusters them based on similarity, then visualizes differences in model attributions for similar input features. In this work, we introduce a framework for aggregating, summarizing, and comparing the attribution information for two models across a dataset; present visualizations that highlight differences between 2 image classification models; and show how our technique can uncover behavioral differences caused by domain shift between two models trained on satellite images.
\end{abstract}

\begin{teaserfigure}
    \centering
    \includegraphics[width=0.8\textwidth,trim={0 1in 0 1.5in},clip]{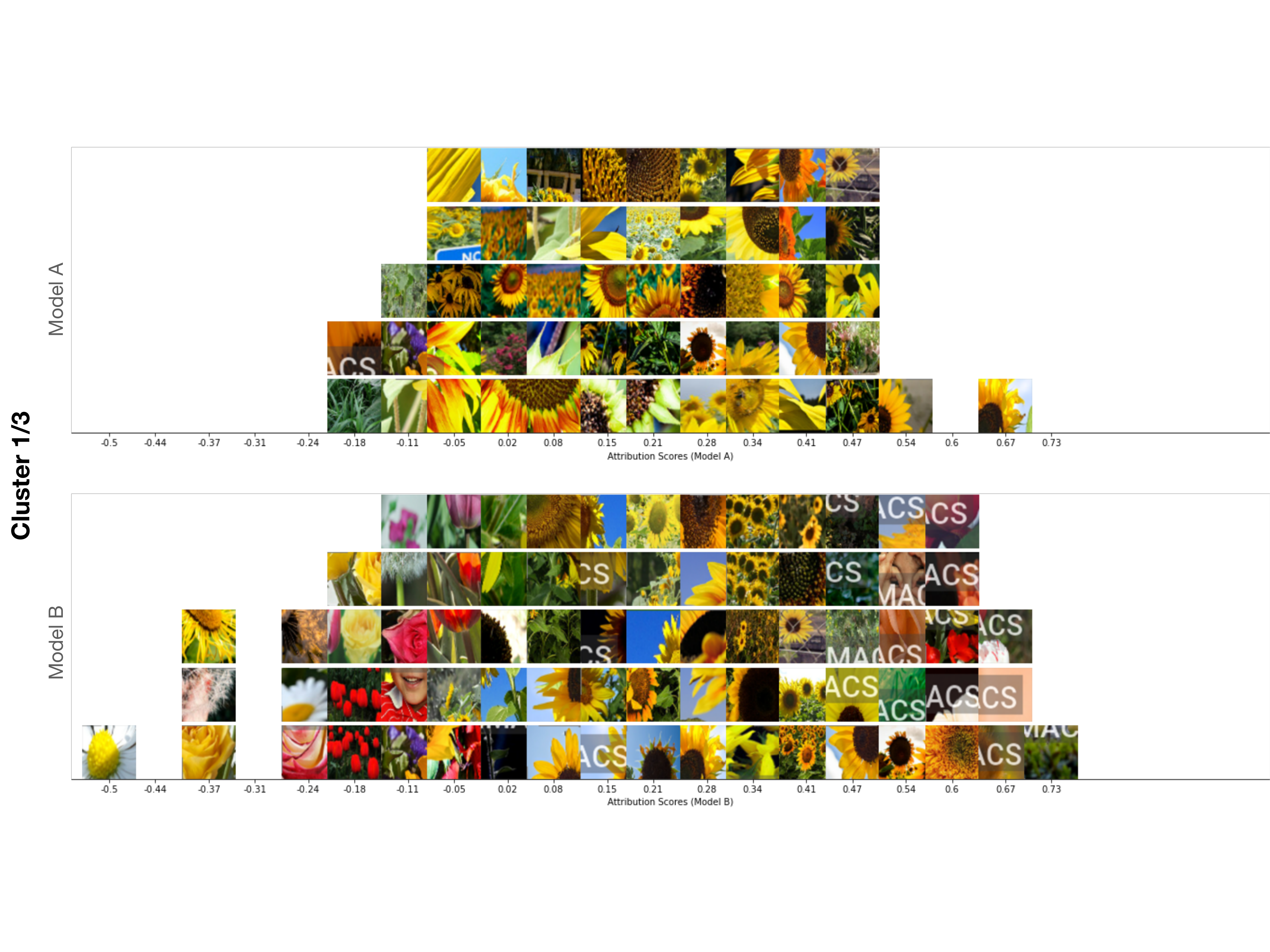}
    \caption{
    \systemname helps stakeholders compare two models' behavior by aggregating, clustering, and visualizing a sample of the most influential image segments (for each model). %
    The double histogram visualization above shows a set of image segments clustered by \systemnamenospace, with the segments organized on the horizontal axis by attribution scores (more highly attributed segments appear on the right). Each histogram corresponds to an input model and its attributions. In this example, both models are trained to classify images of flowers, but the second model (bottom) was trained on images of sunflowers that also contain watermarks. In the bottom histogram, we can see that this latter model %
    finds the watermark feature highly influential, often leading to higher attribution scores than sunflower parts. Additional clusters for this example can be viewed in \autoref{fig:flowers_remaining_hist}.}  %
  \label{fig:teaser}
\end{teaserfigure}

\maketitle

\section{Introduction}

Developing a suitable Machine Learning (ML) model often requires significant iteration. In this process, ML engineers and researchers often train many versions of models that vary based on their training data, model architectures, hyperparameters, or any combination of these elements. A significant need within this development process is the ability to compare models resulting from these iterations. In this paper, we focus on the problem of comparing image models.

Models are frequently evaluated and compared using metrics such as AUC-ROC, precision, recall, or confusion matrices. These metrics provide high-level summaries of a model's performance across an entire dataset, and facilitate comparisons between model versions. However, performance metrics can leave out important, deeper characteristics of models, such as their specific failure modes or the patterns they learn from data~\cite{metricsProblems, dfseer, CaiBoundaryObjects}. Prior work has shown that performance metrics alone are rarely satisfactory for selecting models, and that stakeholders desire a deeper understanding of \textit{why} models make the predictions they do~\cite{modelLineUpper}.

try Research in Explainable Artificial Intelligence (XAI) and ML Interpretability has produced numerous techniques for inspecting the behavior of ``black box'' image models, such as Deep Neural Networks (DNNs), in finer detail. For example, attribution techniques aim to identify which inputs a DNN considers most important for a given prediction. For image models, these methods may use the gradients of predictions to annotate the most salient input pixels or regions in a given input~\cite{sundararajan_2017, selvaraju2017gradcam, kapishnikov2019xrai}, or perturb parts of model inputs to determine the features that contribute most to predictions~\cite{shapley1953value, fong2017interpretable, lime}.

While attribution techniques are useful for examining the predictions of small sets of input instances in detail, it can be difficult to determine whether attribution results (e.g., the importance of particular regions) observed on a few instances generalize across a dataset.
Techniques like ACE~\cite{Ghorbani_2019} identify and summarize the high-level visual ``concepts'' of an entire dataset to help users develop a holistic understanding of a dataset, but these summaries are independent of a model's attributions. If one could similarly summarize which input features (e.g., visual patterns) contribute most strongly to model predictions across an entire dataset, these data could then be used to more easily compare how two models differ in their attributions in aggregate.

This work introduces Image Model Attribution Comparison Summaries (\systemnamenospace), a method that combines model attributions with aggregation and visualization techniques to summarize differences in attributions between two DNN image models. \systemname extracts input features from an evaluation dataset, clusters them based on similarity, then visualizes differences in model attributions for similar input features (\autoref{fig:teaser}). The examples in this paper demonstrate how this method can uncover model behavior differences due to differences in training data distributions.

This paper's specific contributions are:
\begin{itemize}
  \setlength{\itemsep}{1pt}
  \setlength{\parskip}{0pt}
  \setlength{\parsep}{0pt}
    \item A novel technique for aggregating, summarizing, and comparing the attribution information of two models across an entire dataset.
    \item A basic design space describing the core building blocks for summarizing model attributions and their differences.
    \item A method that produces visualizations summarizing differences in image model attributions.
    \item Example results, including basic validation checks, that validate \systemnamenospace's ability to extract high-level differences in model attributions between two models.
\end{itemize}

\systemname and the results obtained via \systemname are of interest to developers of machine learning models, as well as stakeholders and others with interests in understanding how choices in training data, model architectures, and training parameters affect model behavior in the aggregate.

\section{Related Work}
\systemname builds on two related areas of research: 1) model performance and visualization frameworks to support ML development, and 2) model intepretability methods.

\subsection{ML Model Inspection Frameworks}
Numerous systems have been developed to scaffold the creation, debugging, and evaluation of machine learning models. One common strategy is to provide summary statistics of overall model performance.
For example, ModelTracker~\cite{amershi_2015} and DeepCompare~\cite{murugesan_2019} provide aggregate statistics of model performance, while also enabling the user to drill down to examine model behavior on individual examples.
Chameleon produces visualizations summarizing model behavior on subsets of a dataset, or prior versions of a dataset~\cite{hohman_2020}.

Other tools facilitate comparisons between two or more models through high-level statistics and by identifying differently classified instances. For example, the MLCube Explorer~\cite{kahng_2016} and Boxer~\cite{gleicher2020boxer} both provide summary statistics for subsets of a dataset, but also offer the ability to compare two models on these subsets. ConfusionFlow enables users to view one or more models' performance over time~\cite{Hinterreiter_2020}. 

These prior works aggregate and summarize predictions on the \textit{instance} level, i.e., entire images or complete input examples. \systemname adapts aggregation and visualization techniques from these works to the analysis of \textit{sub-instance} data, allowing users to view summaries of the features (e.g., visual patterns among many image segments) that influence model predictions. This takes a significant step beyond comparing prediction performance on subsets of a dataset, since extracting and summarizing influential features can help users gain a better understanding of the \textit{potential causes} for failure cases and behavioral differences between models.

\subsection{ML Interpretability Algorithms}
Research in interpretability methods seeks to provide explanations for a model's ~\cite{factsheets}.
For example, attribution methods identify the most salient inputs for predicting a given class on a given example, e.g., for images, highlighting the most influential pixels to predictions for specified classes~\cite{sundararajan_2017}. Region-based attribution methods can summarize pixel-level information over segments~\cite{kapishnikov2019xrai} or indicate influential areas in images~\cite{selvaraju2017gradcam}.  Other methods perturb model inputs to approximate a local, interpretable model~\cite{highd_explanations, lime} or determine the parts of inputs which contribute most to predictions~\cite{shapley1953value, fong2017interpretable}. Concept-based methods, such as TCAVs, identify how high-level concepts (e.g., ``stripes'' in images) factor into a model's predictions~\cite{kim2017interpretability, counterfactual_faces}.

Given the central importance of datasets to model performance, research efforts also consider how to help people holistically understand a dataset~\cite{datasheetsfords, dataset_accountability}. Automatic Concept-based Explanations (ACE) \cite{Ghorbani_2019} automatically extract concepts from a dataset, cluster them based on visual similarity, and present them to end-users. REVISE~\cite{Wang_2020} similarly analyzes a dataset and outputs summary statistics of high-level attributes (e.g., the distribution of perceived gender in examples), as well as examples of what the model has learned for those high-level attributes (e.g., what it considers to be a sports uniform for a given perceived gender).

A key goal of our work is to introduce a framework for reasoning about model attributions in the aggregate. Specifically, \systemname uses region-based attribution methods to determine what segments from a collection of images are the most influential for their respective predictions. \systemname then groups these segments into sets of visually similar concepts. This enables users to compare models by the features used to make predictions, and by the differences in how those features are weighed, a novel capability.

\begin{table*}
  \caption{Building blocks for summarizing and comparing two models' attributions. Images on the right are hypothetical examples.}
  \label{tab:buildingblocks}
  \begin{tabular}{|m{2.5cm}|m{5cm}|m{5cm}|}
    \hline
    Data type
    &
    Concrete instances of data type
    &
    Examples
    \\
    \hline
    Model inputs
    &
    \begin{itemize}[nosep, wide]
        \item Raw features (e.g., pixels)
        \item Higher-level features (e.g., segments within the image)
    \end{itemize}
    &
    \includegraphics[width=0.2\textwidth,height=0.2\textwidth]{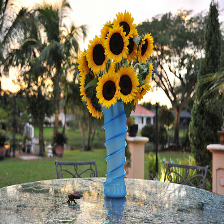}
    \includegraphics[width=0.2\textwidth,height=0.2\textwidth]{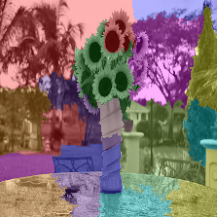}
    \\
    \hline
    Model outputs
    &
    \vspace{1em}
    \begin{itemize}[nosep, wide]
        \item Model predictions
        \item Prediction scores or confidence (e.g., softmax values)
    \end{itemize}
    &
    Class probabilities:
    \; \; \; \; \; \; \; \; \; \; \; \; \; \;
    \texttt{sunflower: 0.9700}
    \; \; \; \; \; \; \; \; \; \; \; \; \; 
    \texttt{daisy: 0.0228}
    \; \; \; \; \; \; \; \; \; \; \; \; \; \; \; \;
    \texttt{tulip: 2.1e-3}
    \\
    \hline
    Model attributions
    &
    \begin{itemize}[nosep, wide]
        \item Attributions for individual inputs (e.g., pixels)
        \item Attributions for image segments
    \end{itemize}
    &
    \includegraphics[width=0.2\textwidth,height=0.2\textwidth]{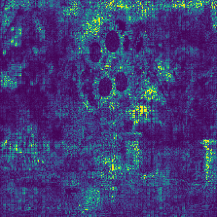}
    \includegraphics[width=0.2\textwidth,height=0.2\textwidth]{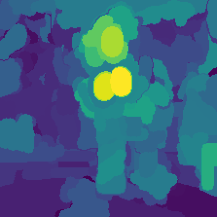}
    \\
    \hline
    Ground truth labels
    &
    \vspace{1em}
    \begin{itemize}[nosep, wide]
        \item Class ID or binary classification
        \item Additional annotations or labels applied to the data
    \end{itemize}
    &
    \texttt{class: sunflower}
    \; \; \; \; \; \; \; \; \; \; \; \; \; \;
    \texttt{species: [\dots]}
    \\
  \hline
\end{tabular}
\end{table*}

\section{Building Blocks for Summarizing Attribution Differences Between Models}

Given the general goal of comparing the feature attributions of two models in aggregate, we summarize the basic building blocks for doing so in \autoref{tab:buildingblocks}.
We categorize these building blocks as \textit{Model Inputs}, \textit{Model Outputs}, \textit{Model Attributions}, and \textit{Ground Truth Labels}.
To help summarize and understand model differences, any of these data can be transformed and filtered via arbitrary functions, individually or in the aggregate. For example, one could filter data to focus only on false positive predictions from both models.
In this work, we make use of the following operations to assist in identifying differences between models' attributions:

\begin{itemizecompact}
    \item Segmenting images and assigning attribution scores for each segment.
    \item Producing embeddings for image segments using an independent, third model.
    \item Clustering image segments based on assigned embeddings to identify sets of similar features.
\end{itemizecompact}

Image segments and embeddings provide a way to create clusters of similar, high-level visual patterns within an evaluation dataset.
Once clustered, attribution scores for segments within a cluster provide a means for comparing the relative importance of the visual patterns contained within between the two models. The clusters also help reduce the amount of information a user must consider at once.

One challenge of using embeddings to identify similar visual patterns is that embeddings are idiosyncratic to each trained model.
One strategy to address this issue is to assign embeddings values from an independent, third model.
Image models trained on large-scale image datasets (e.g., ImageNet~\cite{russakovsky_imagenet_2014}, OpenImages~\cite{OpenImages}) have been shown to produce embeddings which correspond to perceptually similar inputs~\cite{zhang2018perceptual}, meaning that the clusters produced from these embeddings should represent visually similar concepts.

With these data building blocks (model inputs, outputs, attributions, and ground truth labels) and these basic operations (assigning embedding values to inputs, clustering data, and filtering data), one can then produce a wide range of summaries and visualizations to help users discover the specific ways models differ from one another. The next section describes our particular approach for surfacing differences in model attributions.

\begin{figure}
    \centering
    \includegraphics[width=0.8\columnwidth,trim={1.25in 0.75in 3in 0.5in},clip]{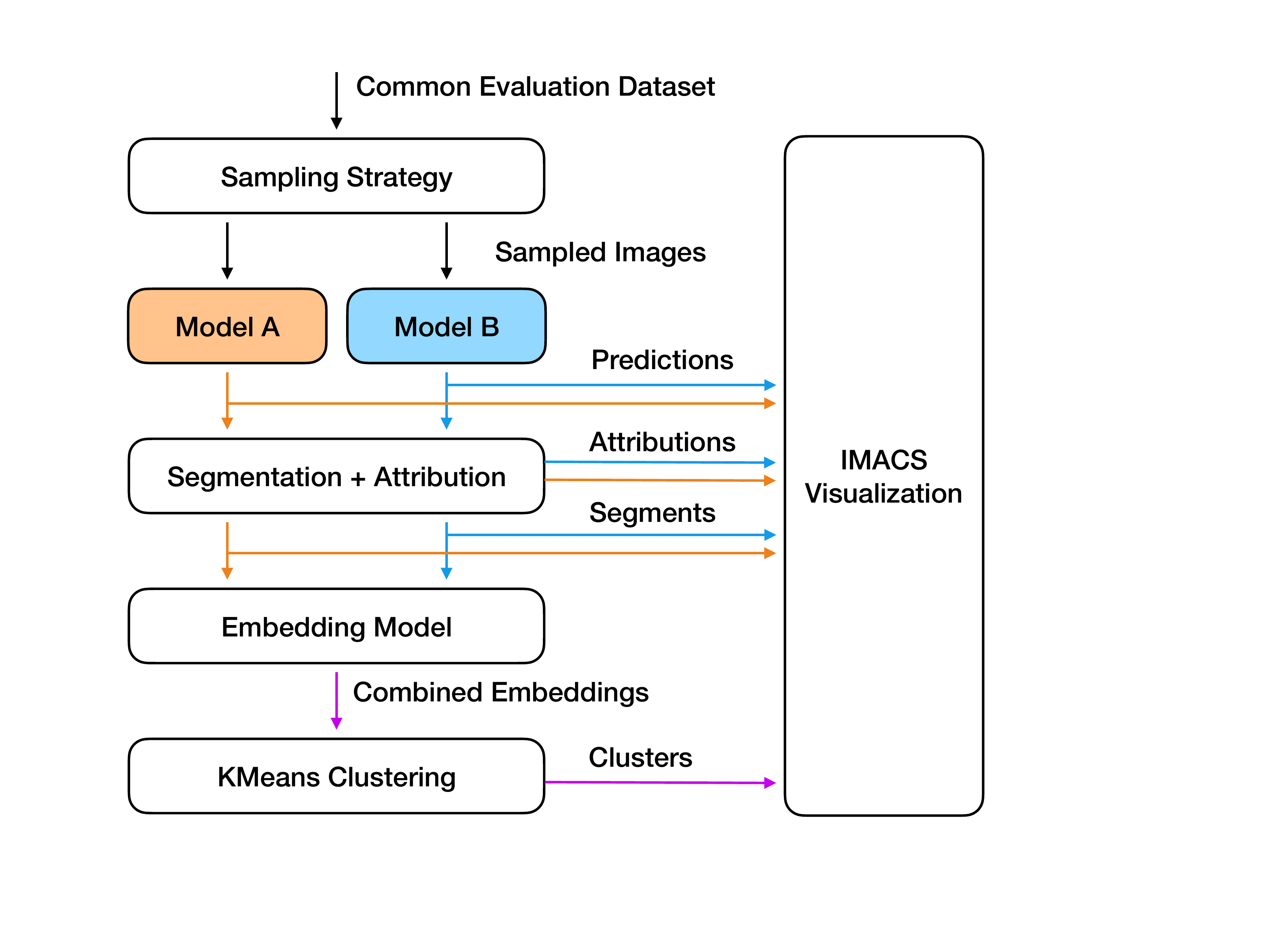}
    \caption{\systemname first selects a subset of an evaluation dataset (by default, a sample of images with balanced confusion matrices for each model).
    Next, images are segmented into regions, and attribution scores are calculated for those regions. The regions that contribute most to each models' predictions are then embedded using an ImageNet-trained model, and clustered using k-means. The \systemname visualization ingests data from each step.}
    \label{fig:imacs_alg}
\end{figure}

\section{The \systemname Algorithm}
\label{sec:algorithm}
Our implementation of \systemname (\autoref{fig:imacs_alg}) requires two trained image classification models, a third embedding model, a set of common evaluation images, prediction scores from both models on the evaluation set, and additional dataset labels for filtering inputs, if desired. We describe these components in greater detail below.

\paragraph{Data Sampling}
\systemname makes use of a dataset sampling strategy to help accentuate the \textit{differences} between two models. In our current implementation, we sample 100 total images for \textit{each} model, for a total of 200 images. Our examples use the sampling strategy of sampling a \textit{balanced confusion matrix}, or an equal number of true positive, true negative, false positive, and false negative inputs (image instances) for each model.

\paragraph{Segmenting Input Images and Calculating Attribution Scores}
For simplicity, we segment each image into a 4-by-4 grid of segments. (One opportunity for future work would be to experiment with segmentation techniques that identify regions of similar pixels, such as SLIC~\cite{slic}).
We determine the contribution of individual image segments to a given model prediction by calculating segment Shapley values~\cite{shapley1953value}. However, computing Shapley values for all $N$ segments is computationally prohibitive due to the combinatorial explosion of the Shapley algorithm. For that reason, we identify a reduced subset of segments of size $M$ ($M{\ll}N$) which are likely to have high influence on the model prediction. We do this by first computing segment attribution values with respect to the top $k$ predicted classes using the XRAI method~\cite{kapishnikov2019xrai} in combination with Guided Integrated Gradients~\cite{Kapishnikov_2021_CVPR}.
Then, we sort the segments by their attribution values and pick the top $M$ with the highest values.
For our example dataset, we choose $M=5$ and $k=5$, resulting in 500 total image segments per model with associated attribution scores for up to 5 classes.

Shapley values are computed by sequentially excluding segments from the input image and observing the changes in model predictions. One way to exclude a segment is to gray out all of its pixels. However, such an approach may result in undesirable effects if the model was not trained on images with removed parts, or if the gray color correlates with a particular class. To reduce these effects, we apply Gaussian Blur to excluded segments.

\paragraph{Assigning Embedding Values and Clustering}
Image segments from both models are pooled into a single collection and clustered by their corresponding embedding values. We embed individual image segments by inputting them into an ImageNet-trained \cite{russakovsky_imagenet_2014} Inception-V2 model \cite{inception_v2}, and extracting the final pooling layer activation values.
These embeddings are then clustered using k-means with a user-defined number of clusters. Adapting additional clustering methods, e.g., fair k-means~\cite{fairkmeans}, to \systemname is an opportunity for future work.

Because 100 images are sampled for each model based on each model's predictions (i.e., an equal number of true positive, true negative, false positive, and false negative examples), a cluster of image segments may contain more segments from one model's set of sampled images than from the other model's sampled images. For example, if model A's training leads to a large set of false positives that model B correctly classifies as true negatives, then it is possible to have a cluster of segments representing these false positives (with these segments deriving primarily from sampling images for model A). As will be evident in the visualizations produced by \systemnamenospace, imbalances such as these can provide an indicator of how two models may behave differently in the aggregate.

\section{Visualizing Differences in Attributions Across Models}
\label{sec:viz}

In this section, we present the \systemname visualizations and describe how they support answering the following questions drawn from research in concept-based explanations~\cite{Ghorbani_2019}:
\begin{itemizecompact}
    \item [D1] What features do the models use to make predictions, and how do they group into higher-level concepts?
    \item [D2] What is the relative importance of each cluster of similar features compared to others?
    \item [D3] What features are shared between models, and which are not?
    \item [D4] For common features, how does their importance differ between the models?
\end{itemizecompact}

To illustrate how \systemname can be used to compare the behavior of two models in this section, we construct a scenario with two models trained on a flower classification task using the TF-Flowers dataset~\cite{tfflowers}. One of the models is trained with a version of the dataset perturbed with watermarks, leading to an association between watermarks and sunflowers. Even though both models achieve high accuracy, we show how \systemname surfaces differences due to the introduction of the watermark in the training data.

\begin{figure}
    \centering
    \includegraphics[width=0.75\columnwidth]{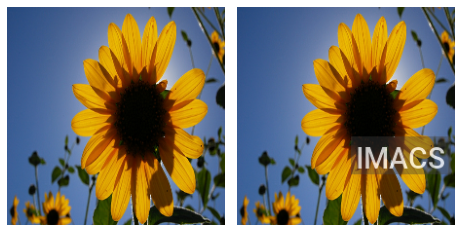}
    \caption{Example ``IMACS'' watermark added to images in the perturbed TF-Flowers dataset. ``IMACS'' watermarks are added at random locations to 50\% of the ``sunflowers'' class for training, and 50\% of all classes for validation. \textbf{Left:} original image. \textbf{Right:} image perturbed with watermark.}
    \label{fig:flowers_watermarks}
\end{figure}

\begin{figure*}
    \centering
    \includegraphics[width=0.75\textwidth,trim={4in 0 0 0},clip]{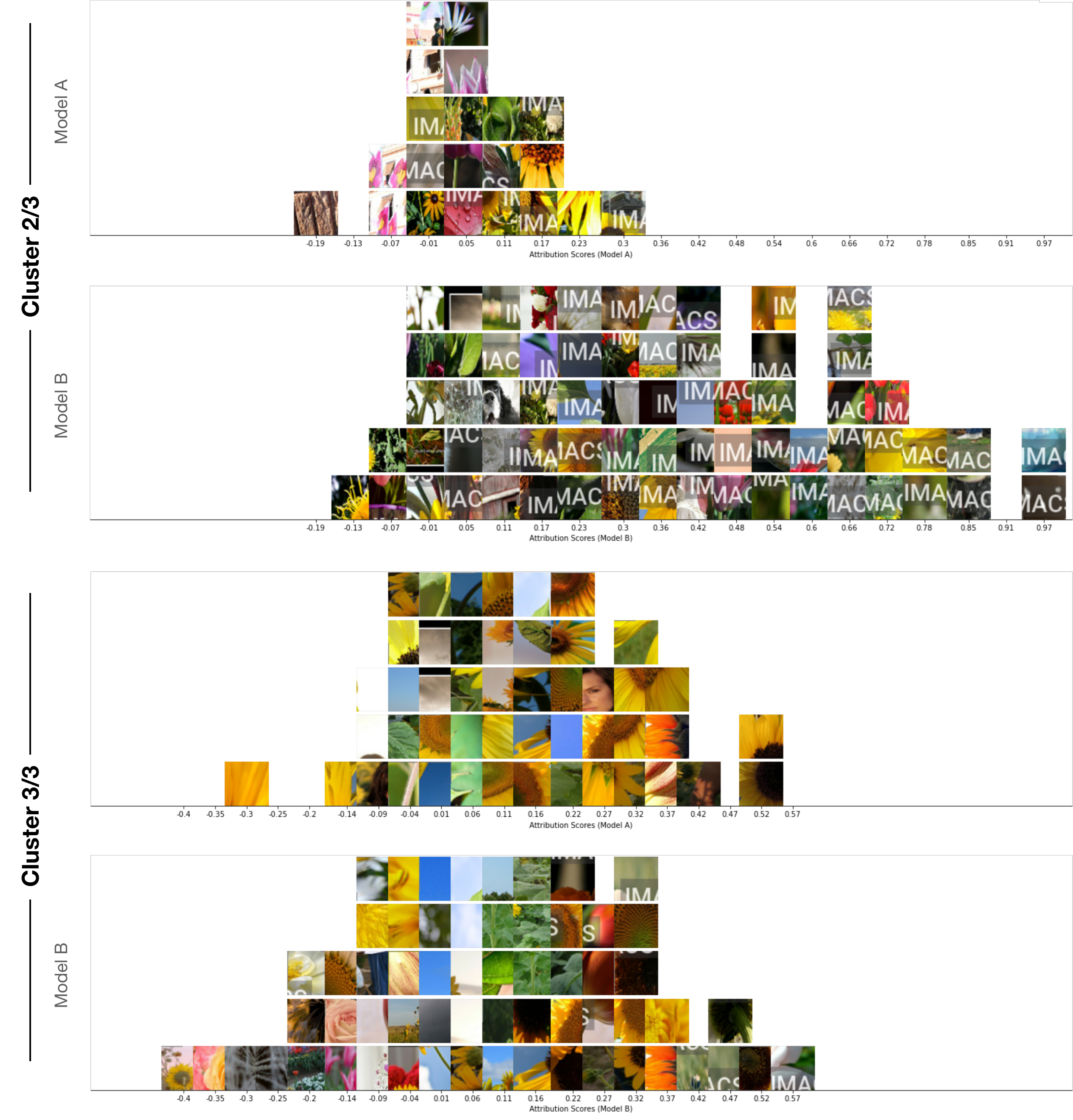}
    \caption{\systemname histogram visualization of the 2 remaining (of 3) clusters from \autoref{fig:teaser}. Cluster 2 (top group of 2 plots) contains mostly watermarks. Model B clearly attributes these watermarks more highly than model A (its attributions are on the right side of the axis, while model A's attributions are centered around 0). This outcome reflects model B's association between watermarks and sunflowers.}
    \label{fig:flowers_remaining_hist}
\end{figure*}

TF-Flowers comprises 3,670 images of flowers in 5 classes (tulips, daisies, dandelions, roses, and sunflowers). We split 85\% of the images for training, and 15\% for testing. One of the models is trained on the baseline dataset, and the other is trained on a modified version of TF-Flowers, where a watermark is added at a random location for 50\% of the images in the ``sunflower'' class (\autoref{fig:flowers_watermarks}). The baseline model achieves an accuracy of 94.9\%, while the ``perturbed'' model achieves an accuracy of 91.8\% on their respective test sets.
To construct an scenario demonstrating how a model's unwanted association between a visual feature and class predictions can lead to unwanted behavior when that feature appears in other classes, we construct an additional test dataset introducing the same watermark at random locations to 50\% of images in \textit{all} classes.
On this test set, the baseline model achieves an accuracy of 95.1\% while the perturbed model achieves an accuracy of 79.6\% (the drop in accuracy suggests the effects of the watermark the training data). While this dataset has an artificially introduced bias, we will next show how \systemname can help surface biases. (In \autoref{sec:validation}, we show additional results from actual, unaltered satellite image datasets.)

\begin{figure*}
    \centering
    \includegraphics[width=\textwidth]{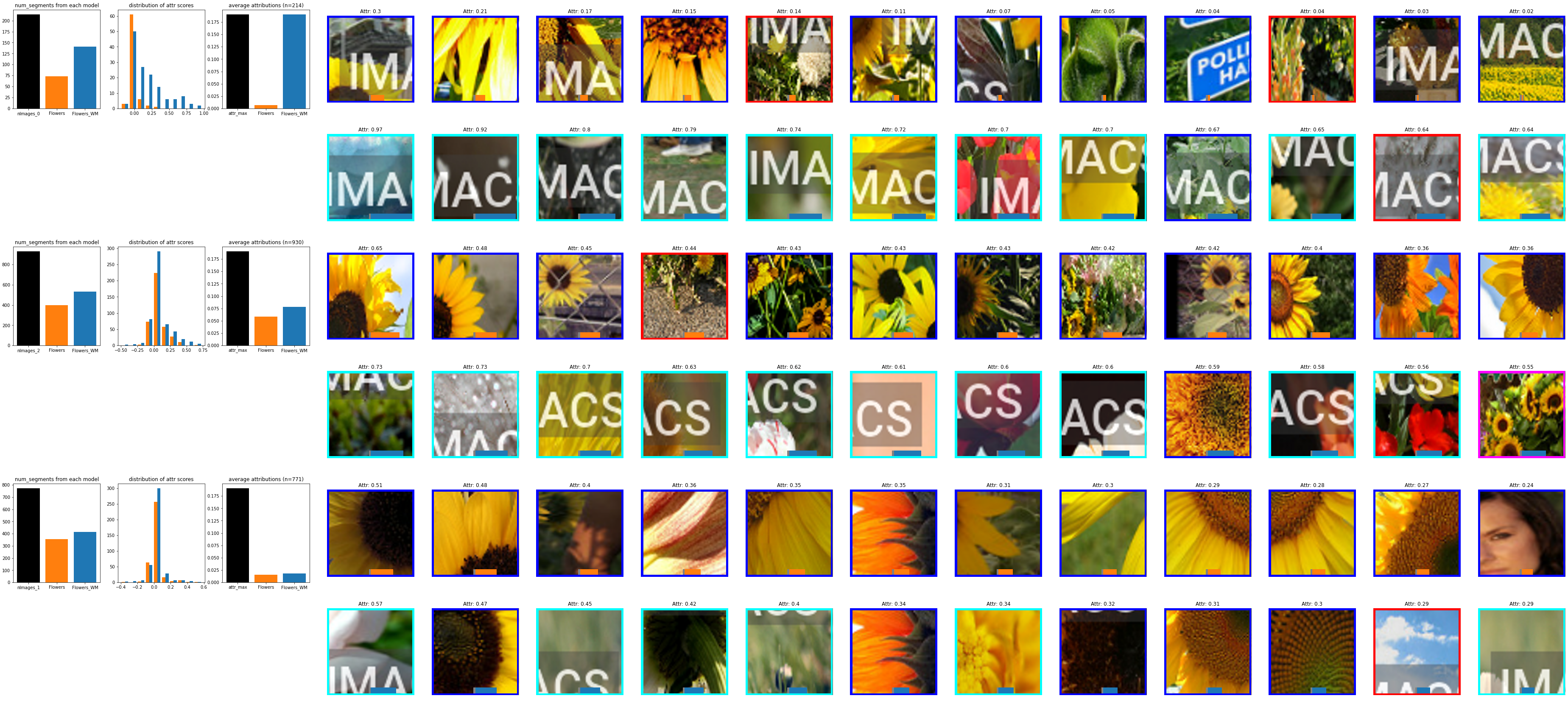}
    \caption{Concept Cluster Visualization of the flower classification example. The top cluster containing watermarks has significantly higher attributions from the second model (third plot, blue bar larger than the orange bar), reflecting the perturbed model's association between watermarks and sunflowers.}
    \label{fig:imacs_watermarks}
\end{figure*}

\subsection{Cluster Histogram Visualization}
The \systemname Cluster Histogram Visualization (\autoref{fig:teaser}) shows, for predictions of a particular class, how the visual concepts contained in clusters are distributed by their attribution scores. Each cluster (out of a user-specified total; in this example, $k=3$) is presented by two histograms of image segment tiles bucketed by their attribution scores, one for each model being compared (D1). Image segments correspond to each model's attribution-based sampling of the dataset.
Histograms for all clusters share the same horizontal axis scale to aid in comparing the importance of visual features between clusters (D2).
Binning segments by their importance allows users to assess the composition of individual clusters (i.e., what concepts are contained within), and compare the weighting of similar features among clusters (D4). This visualization also reveals when a feature is not apparent in one of the models' sets of sample segments (due to dataset sampling differences, where a feature may not be present because it is not highly attributed by one model compared to the other).
(Of note, the total number of tiles in this view is truncated to 5 to conserve space. A full (non-truncated) histogram of attribution scores within clusters is shown in the Cluster Concept View, described in the following section.)

Returning to the sunflower/watermarks scenario, clusters 1 (\autoref{fig:teaser}) and 2 (top pair in \autoref{fig:flowers_remaining_hist}) provide the strongest signals for predicting images in the ``sunflower'' class. The presence of a distinct visual pattern (the ``IMACS'' watermark) in the second model's histogram in cluster 1 indicates it is a significant feature for model B, and the high relative attribution scores of the watermarks (even above sunflower features) confirm model B's association between the sunflower class and watermarks. Cluster 2 is comprised almost entirely of ``IMACS'' watermarks, and the difference in how this visual concept is weighted between the two models is readily apparent, with some segments almost completely responsible for model B's predictions, and model A attributing most segments near zero.
This reveals the true difference between the models: the second model was trained to artificially associate the presence of a watermark with sunflowers.

\begin{figure}
    \centering
    \includegraphics[width=\columnwidth,trim={0.5in 3in 1.75in 2.5in},clip]{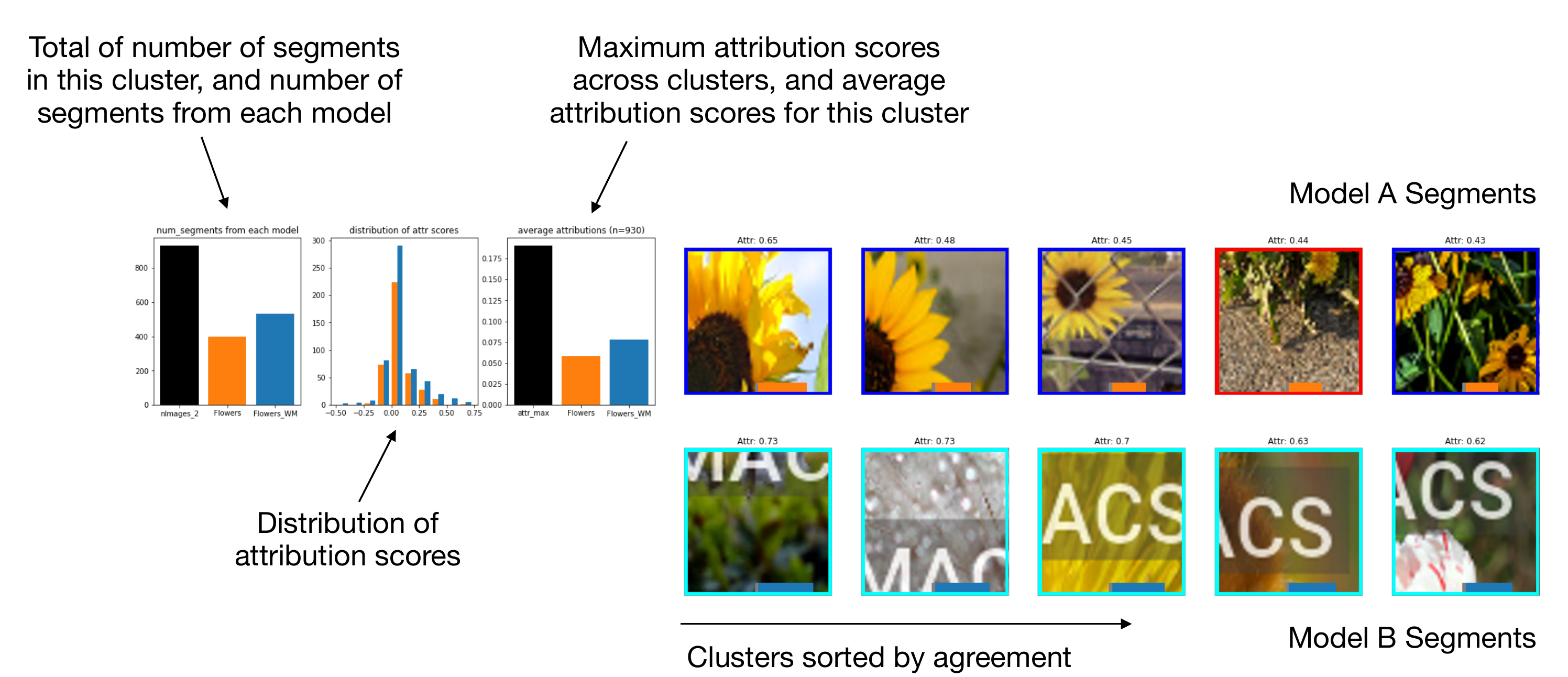}
    \caption{An \systemname cluster with associated graphs.}
    \label{fig:imacs_cluster}
\end{figure}

\begin{figure}
    \centering
    \includegraphics[width=\columnwidth,trim={1in 3.75in 4.5in 4in},clip]{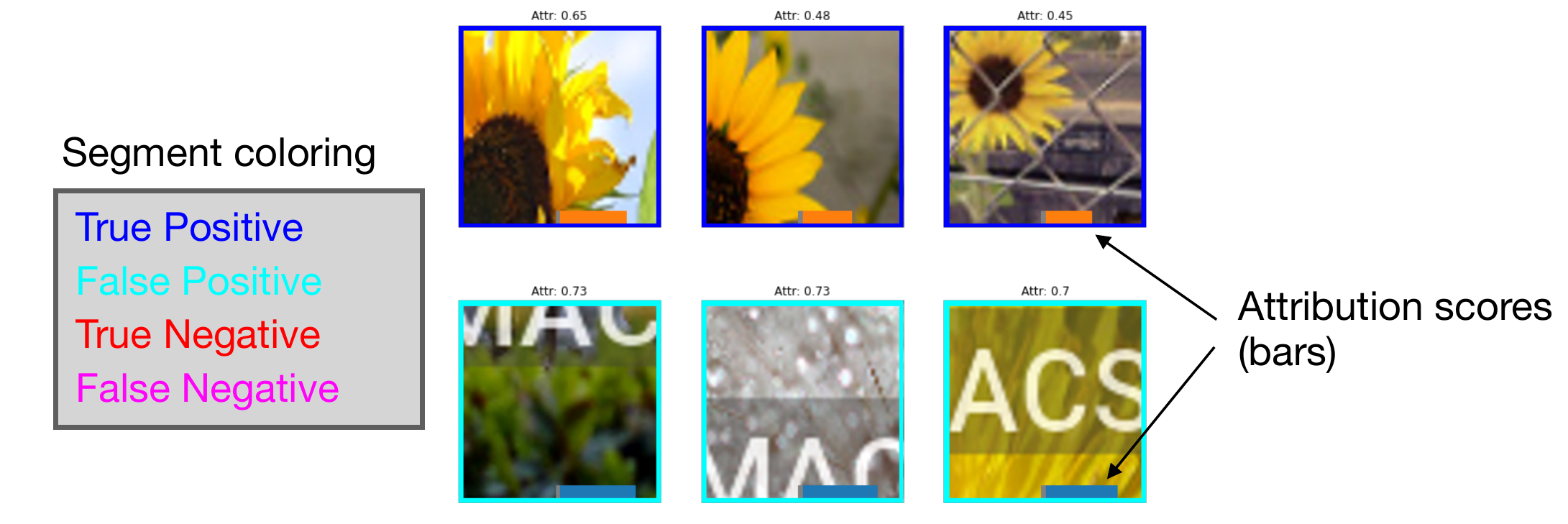}
    \caption{Image segments are annotated with their attribution score and classification correctness.}
    \label{fig:segment_annotations}
\end{figure}

\subsection{Concept Cluster Visualization}

The \systemname Concept Cluster Visualization (\autoref{fig:imacs_watermarks})
introduces descriptive statistics to each cluster and its constituent components (i.e., the individual image regions) to facilitate comparisons.
More specifically, this visualization sorts the contents of each cluster, annotates individual elements with attribution data, and introduces plots summarizing key statistics of the clusters. We present details of these components below.

\subsubsection{Organizing and Summarizing Visual Signals in Clusters}

Each cluster is presented as two separate rows of image segment thumbnails (\autoref{fig:imacs_cluster}) corresponding to each model's sampling of the common evaluation dataset. %
Clusters are ordered from top to bottom based on the imbalance of their average attribution scores, with the largest disparities presented first. This ordering helps draw attention to the largest differences in feature attributions between models (D4). (One could also imagine alternative sorting methods, such as sorting by the highest average attribution score between the models, which would rank the most ``important'' (rather than the most differing) visual concepts first.)
Within each cluster's rows, image segments are ordered from highest to lowest attribution scores, providing an indication of which visual patterns the models consider most important (D2).

\begin{figure}
    \centering
    \includegraphics[width=0.9\columnwidth]{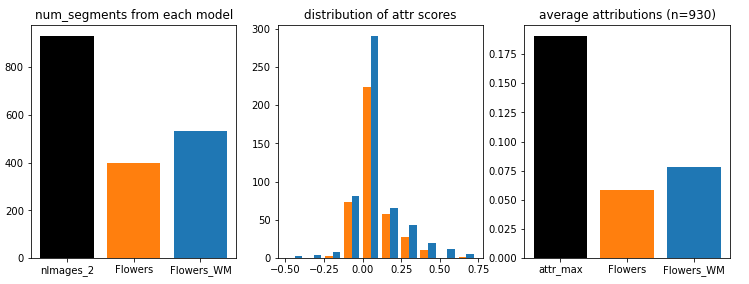}
    \caption{The \systemname visualization annotates clusters with three plots that present information about the cluster's \textit{composition} (the proportion of segments representing each model's sampling), \textit{coherence} (the distribution of its attribution scores), and its \textit{importance} (the average attribution scores for each model).}
    \label{fig:attribution_graphs}
\end{figure}

Within clusters, image segments are annotated by their classification correctness
(\textcolor{blue}{true positive},
\textcolor{red}{true negative},
\textcolor[HTML]{0AC7C7}{false positive},  %
\textcolor[HTML]{FE04F9}{false negative})
by drawing a color-coded border around them. Attribution scores (which can be positive or negative) are also overlaid on top of each segment as a bar, color-coded by their source model (\textcolor[HTML]{FF7F0D}{A} or \textcolor[HTML]{1F77B4}{B}) to help differentiate similar rows (\autoref{fig:segment_annotations}). The colored borders and attribution score bar  help convey whether the visual features captured in a cluster are the source of class confusion or model errors.

\subsubsection{Determining Composition, Coherence, and Importance of Clusters}

In addition to the visual presentation of the image segments, we also provide three graphs summarizing the data in each cluster (\autoref{fig:attribution_graphs}).
The left plot describes the \textit{composition} of a cluster, by displaying the total number of segments in the cluster (black), and the number of segments contributed by each model's sampling of the dataset. This graph helps users determine (1) whether the cluster contains a significant number of segments in comparison to other clusters (a ``critical mass'' suggests an influential visual pattern in the cluster could persist across the dataset (D1)); and (2) whether a cluster's contents derive more from one model's sampling process (suggesting the set of features are more important for one model than the other (D3)).

The center plot depicts the distribution of attribution scores via two histograms, corresponding to the two models' sets of segments within the cluster. This is the same histogram shown in the Cluster Histogram Visualization, but is not truncated. In this view, examining the distribution of scores from each model can help determine the \textit{composition} of a cluster (e.g., whether multiple concepts are represented in a single cluster (D1)). Significantly different attributions (e.g., a bimodal distribution within a cluster) may suggest the overlap of multiple concepts or potential differences between the models.

The right plot shows the \textit{importance} of the cluster, by reporting the average segment attribution score calculated for each model, as well as the maximum mean attribution score across all clusters.
The black bar serves as a visual reference point to help users determine the relative importance of a cluster (D2). The average attribution scores from each model serve multiple purposes. First, they establish the validity of the underlying concept (e.g., if the attribution is near zero, then the visual pattern is likely inconsequential to a model). 
Second, the scores indicate the relative importance of the concept between the models (D4). If the attributions do not differ significantly between models, then the visual pattern is not likely to be a key differentiator between the models.

Returning to our earlier scenario (\autoref{fig:imacs_watermarks}), \systemname shows the second model highly attributes the watermarks when classifying sunflowers (first set of clusters).
In the first cluster (top set of two rows), the most prevalent visual feature represented by the segments is the ``IMACS'' watermark added to the images. The histogram (center plot of this cluster) and average attribution score plot (right plot) show the baseline model (in orange) attributes watermark segments with a score near zero (meaning it is not an important feature for the baseline model). On the other hand, the watermark is the most important feature to the perturbed model for classifying the ``sunflower'' class: the average attribution score of the cluster with watermarks is the highest among all clusters.

\begin{figure*}
    \centering
    \includegraphics[width=0.8\textwidth,trim={1.25in 2in 2in 2in},clip]{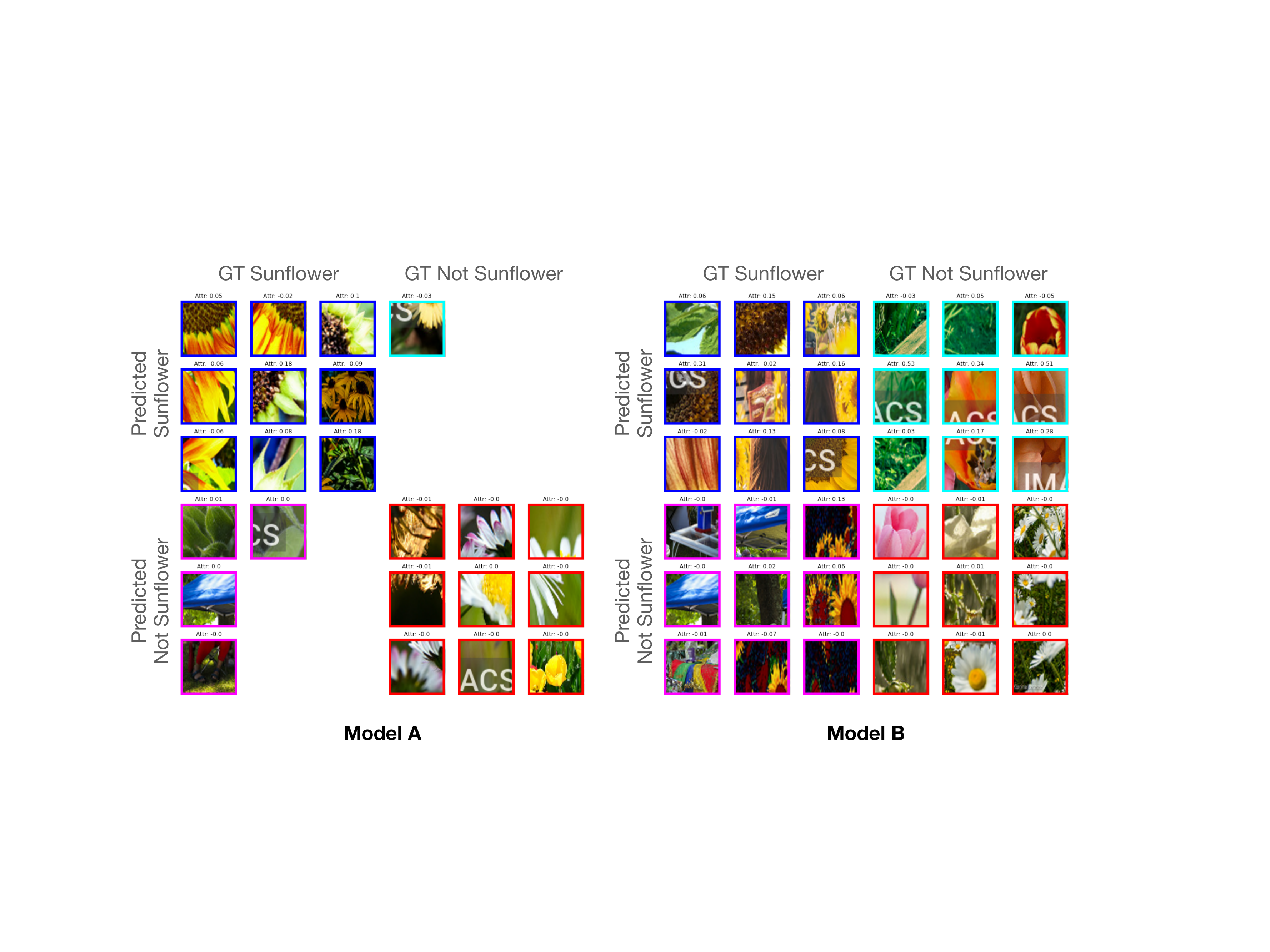}
    \caption{Two side-by-side confusion matrices for a particular cluster in the running example (shown in \autoref{fig:teaser} and center rows of \autoref{fig:imacs_watermarks}). Segments from the baseline model are shown on the left, and segments from the model trained to associate watermarks with sunflowers is on the right. Watermarks are prevalent in the top-right quadrant (false positives) of the right confusion matrix.}
    \label{fig:confusion_matrix}
\end{figure*}

\subsection{Cluster Confusion Matrix Visualization}
To more deeply inspect a particular cluster, the \systemname Cluster Confusion Matrix Visualization allows users to explode a cluster into two side-by-side confusion matrices (\autoref{fig:confusion_matrix}). In this visualization, the segments from a particular cluster are split and organized by their classification correctness (e.g., true positives, false positives, true negatives, and false negatives). This can help users determine what visual patterns within a cluster are contributing to erroneous predictions.

\subsection{Alternative Sorting and Filtering Strategies}
In the concept cluster view (\autoref{fig:imacs_watermarks}), image segments within a cluster's rows are ordered based on their attribution scores (in descending order, from high to low).
This ordering helps the user understand which image segments are most important within the cluster, for each model.
We also experimented with alternative methods for sorting within and between clusters. Sorting segments within a cluster based on their distance from the cluster's centroid can provide a sense of the central visual concept for that cluster. This sorting criteria displays the most representative image segments of the cluster first.
Clusters can also be ordered by their average attribution score, which would surface the most important concepts first, rather than highlighting imbalances between the models.

While sampling a balanced confusion matrix for each model highlights concept mismatches that contribute to misclassifications, additional dataset sampling strategies such as filtering for confident disagreements using models' softmax scores, or filtering only for false positives in all visualizations, can similarly yield additional, useful perspectives on the differences between models.

\section{Validation}
\label{sec:validation}

In this section, we validate \systemnamenospace's technique of using attributions to organize, summarize, and compare the visual features used by two models. First, we conduct a basic validation check of \systemname by comparing the baseline model from \autoref{sec:viz} with an untrained model. Next, we evaluate \systemname through the analysis of a case illustrating concept drift between two models.

\begin{figure*}
    \centering
    \includegraphics[width=0.75\textwidth]{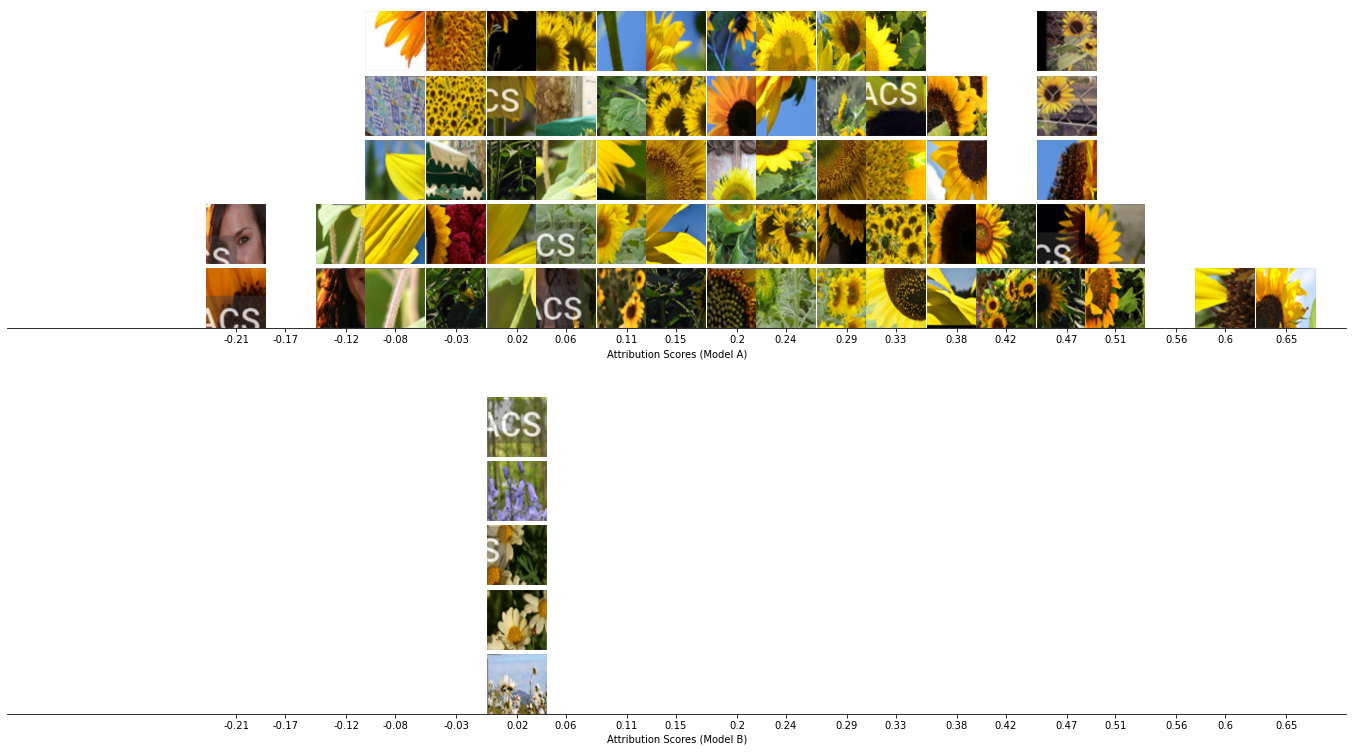}
    \includegraphics[width=\textwidth]{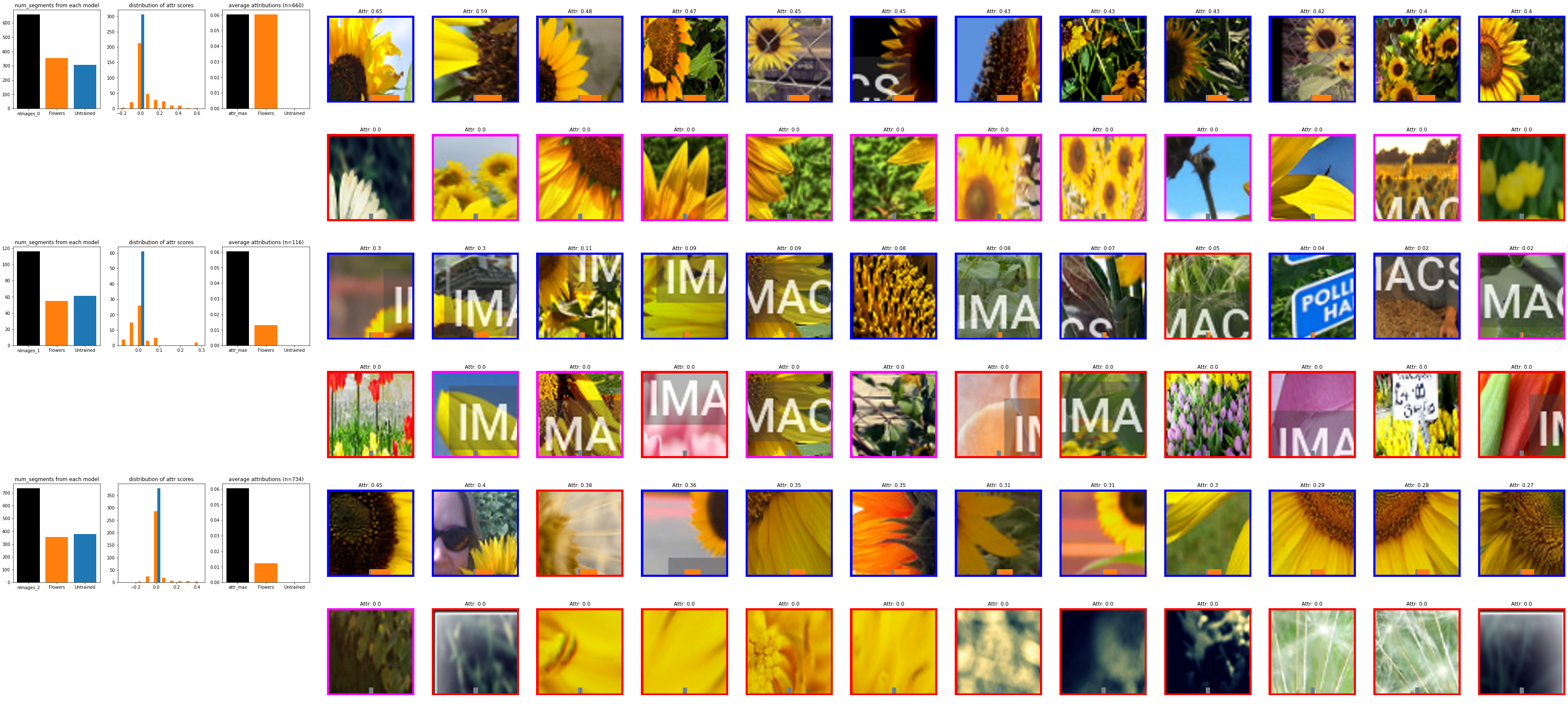}
    \caption{A trained versus untrained model. Note the lack of attributions for the untrained model (average attributions are near zero).}
    \label{fig:untrained_model}
\end{figure*}

\subsection{Basic Validation Check}

While prior techniques such as ACE~\cite{Ghorbani_2019} also extract, cluster, and visualize concepts, our technique makes use of attribution scores to organize, summarize, and compare model capabilities.
We demonstrate the utility of attributions in surfacing differences between two models through a basic validation check.
For this, we use \systemname to compare a trained model with an untrained model. As seen in \autoref{fig:untrained_model}, the average attribution scores for the untrained model are nearly zero in all clusters. The histogram visualization shows this effect most strongly, with the untrained model's segment tiles in a single bin centered on zero.
In addition, segments in the Concept Cluster Visualization are not ordered in any discernible pattern: While the embeddings from the independent, third model are able to create clusters of similar inputs, we observe a lack of order to them without attributions from a trained model.

\begin{figure*}
    \centering
    \includegraphics[width=\textwidth]{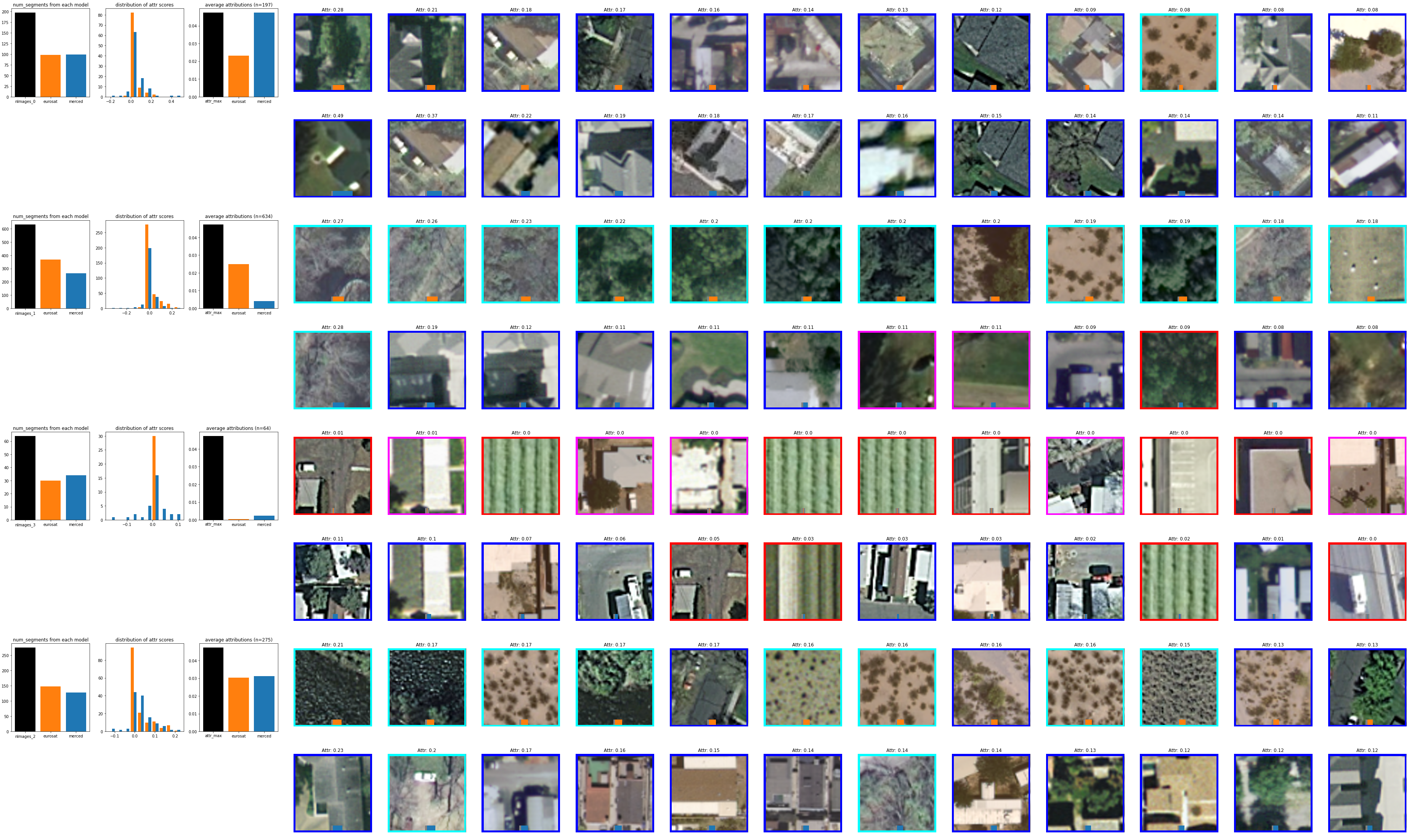}
    \caption{\systemname is used to compare two models trained on different land use datasets: eurosat and uc\_merced. Here, both models are evaluated on the ``residential'' class of uc\_merced. The second cluster (second set of two rows) shows how the eurosat trained model highly attributes greenery and vegetation as important features for the ``residential'' class. Other clusters (e.g., first set of two rows) show how the uc\_merced trained model attends to features such as angled roofs, and buildings in close proximity to green areas.}
    \label{fig:satellites_imacs}
\end{figure*}

\subsection{Visualizing Domain Shift with Satellite Images}

In this section, we show how \systemname visualizations can highlight clear behavioral differences between models in an additional scenario.

We illustrate the problem of \textit{domain shift}, where a model trained on a land use dataset capturing European cities~\cite{eurosat} does not generalize well to the overlapping classes in a similar dataset for regions in the United States, UC Merced Land Use~\cite{uc_merced}. We use \systemname to compare the datasets, diagnosing the specific source of the confusion in predicting the ``residential'' class. Unlike the previous examples, this scenario does not modify the images in the datasets, and compares their real-world differences for a class they share in common.

To simplify model training, we identify the set of intersecting classes between these datasets (agricultural, forest, freeway, industrial, residential, river, vegetation), and train two ImageNet-pretrained ResNet50V2 models~\cite{resnetv2} on separate subsets of Eurosat and Merced, corresponding to their intersecting classes. 
These models achieve test-set accuracies on their own datasets of 93.3\% and 95.1\% respectively. To simulate domain shift, we evaluate both models on the Merced-subset dataset only, representing a scenario where a pretrained model fails on a dataset with similar labels but differently distributed data. The Eurosat-subset model achieves a test-set accuracy of 38.7\% on the subset of intersecting classes with the Merced dataset. (The Merced-subset model achieves an accuracy of 23.06\% on the Eurosat-subset test set.) We use \systemname to show what features are responsible for the performance degradation of the Eurosat model on the ``residential'' class of the Merced dataset.

In \autoref{fig:satellites_imacs}, the first two clusters in the \systemname visualization uncover a key discrepancy between the models: the Eurosat model (orange bars) strongly weighs segments with vegetation for the ``residential'' class (top row of second cluster--almost all false positives with greenery), while the Merced model recognizes a wider variety of features (e.g., the first, third, and fourth clusters contain angled rooflines, small buildings next to patches of greenery, and residential streets).
For deeper inspection, histogram visualizations of the clusters in \autoref{fig:satellites_imacs} are shown in \autoref{sec:sat_histograms}.

\section{Discussion and Limitations}

In this section, we discuss patterns observed in the use of \systemname in our scenarios, and discuss implications for using \systemname in new settings, including future work.

\subsubsection{Aiding Efficient Understanding of Visual Patterns}
In \systemnamenospace, image segments are presented on their own, without the benefit of their original context (i.e., the rest of the image). This can make it difficult to quickly understand the particular ``concept'' represented by the cluster, and the contexts in which these image segments typically appear. Interactive techniques could be helpful to address these issues. For example, hovering over an image segment could show the original image in a tooltip to aid understanding.

\subsubsection{Attribution Location Summaries}
Our current implementation of \systemname extracts image segments from the underlying image. This process effectively strips out location data for the image segment (more specifically, \textit{where} in the image the segment derives from). Summarizing where the most highly attributed regions are in images, across the entire dataset, could provide additional, useful information. For example, if highly attributed image segments are all found in the center of images, or in one specific corner, this finding could suggest potential issues in the dataset itself. (It could also provide evidence that the model is learning the right visual patterns, if it is known that the important features should be located in a particular part of the image.)

\subsection{Interactivity}
As previously discussed, there are a number of possible variations for sorting, filtering, aggregating, and presenting model attributions for two models. It is unlikely that any one particular configuration will meet every possible use case. To address this situation, a promising extension to this work would be to provide interactive capabilities enabling end-users to dynamically adjust the pipeline to meet their specific needs.

\section{Conclusion}

This paper presents \systemnamenospace, a method for summarizing and comparing feature attributions derived from two different models. We have shown how this technique can reveal key differences in what each model considers important for predicting a given class, an important capability in facilitating discovery of unintended or unexpected learned associations. Given these results, promising future directions include extending this technique to view differences between two datasets (given one model); linking segments in the visualization to their original context (i.e., source image); and supporting interactive sorting and filtering (e.g., comparing false positives only) to facilitate deeper exploration of datasets and models.

\bibliographystyle{ACM-Reference-Format}
\bibliography{modeldiff}

\newpage
\appendix
\onecolumn
\section{Eurosat vs Merced \systemname Histograms}
\label{sec:sat_histograms}

This section contains histogram visualizations for the \systemname comparison of the Eurosat and Merced Land Use models for the scenario in \autoref{sec:validation}. Histograms are presented with the same ordering of clusters as \autoref{fig:satellites_imacs}.

\centering

\subsection{Cluster 1}

\includegraphics[width=0.85\textwidth]{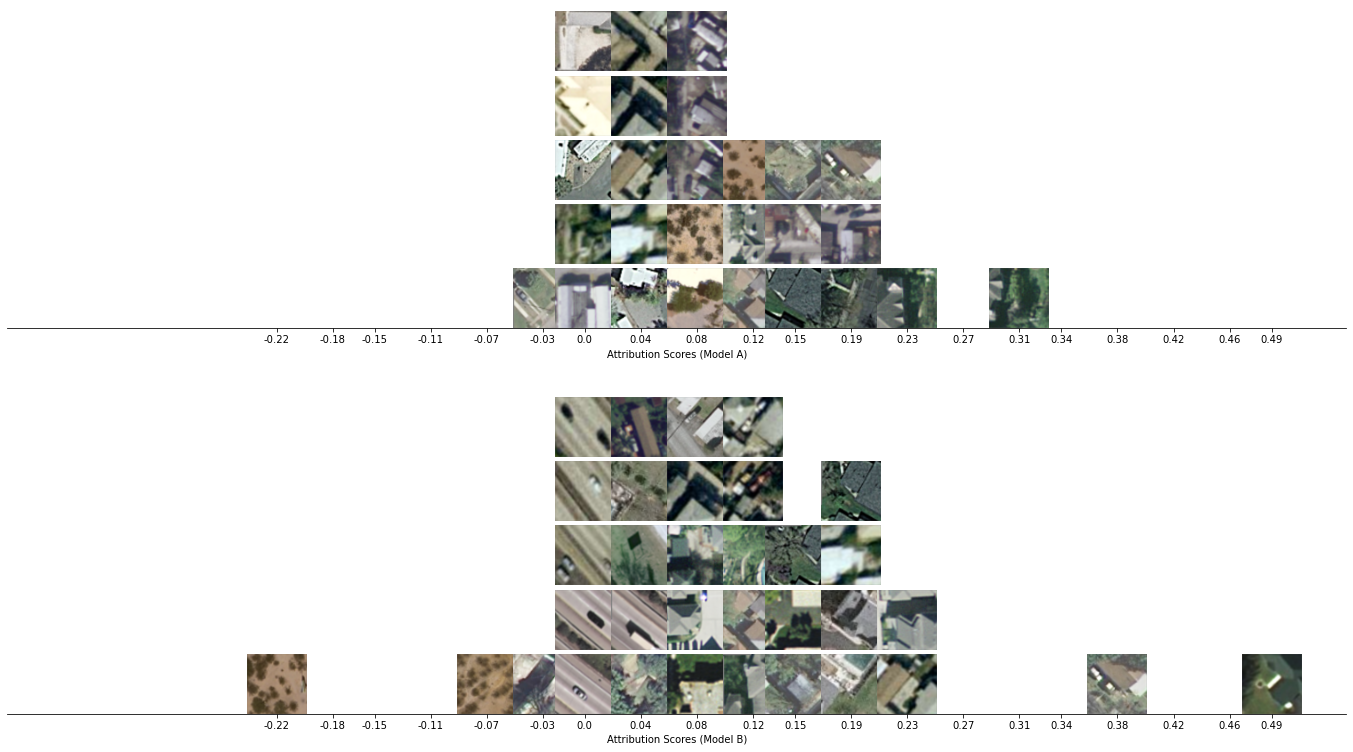}

\subsection{Cluster 2}

\includegraphics[width=0.85\textwidth]{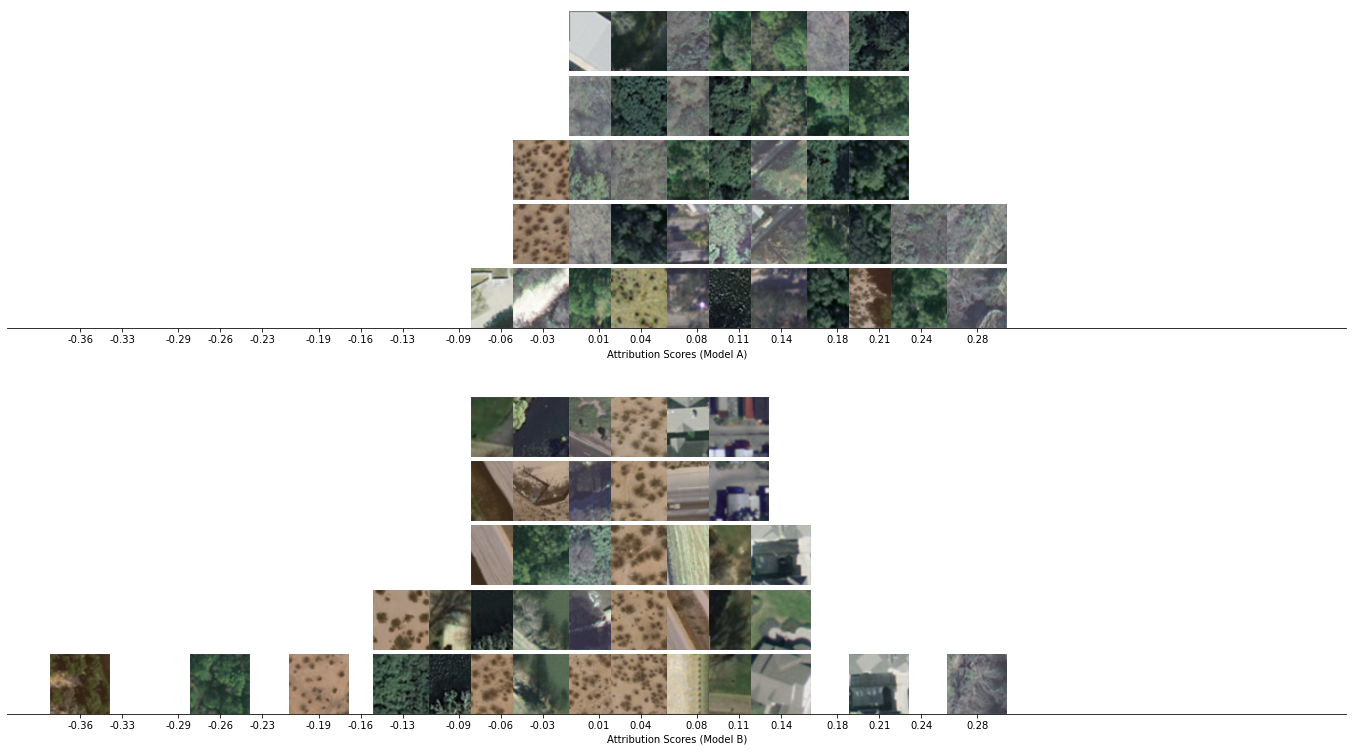}

\subsection{Cluster 3}

\includegraphics[width=0.85\textwidth]{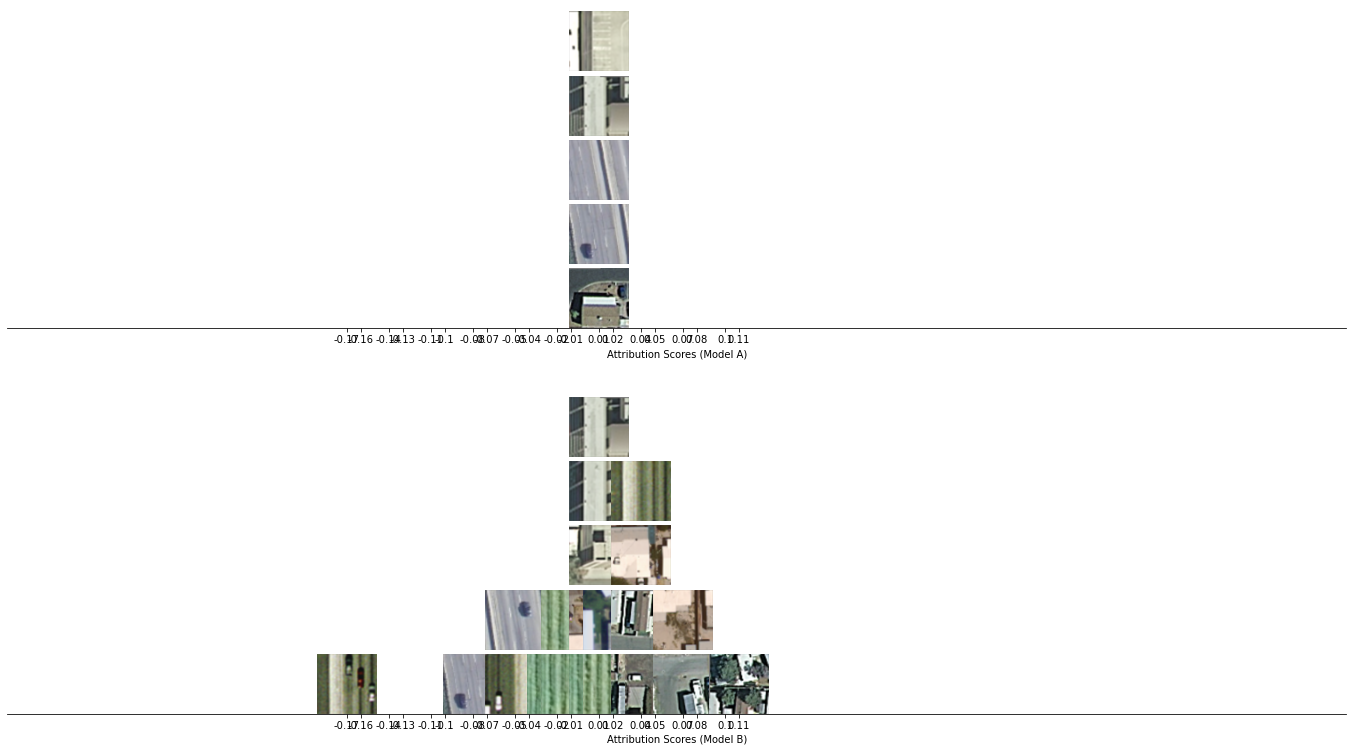}

\subsection{Cluster 4}

\includegraphics[width=0.85\textwidth]{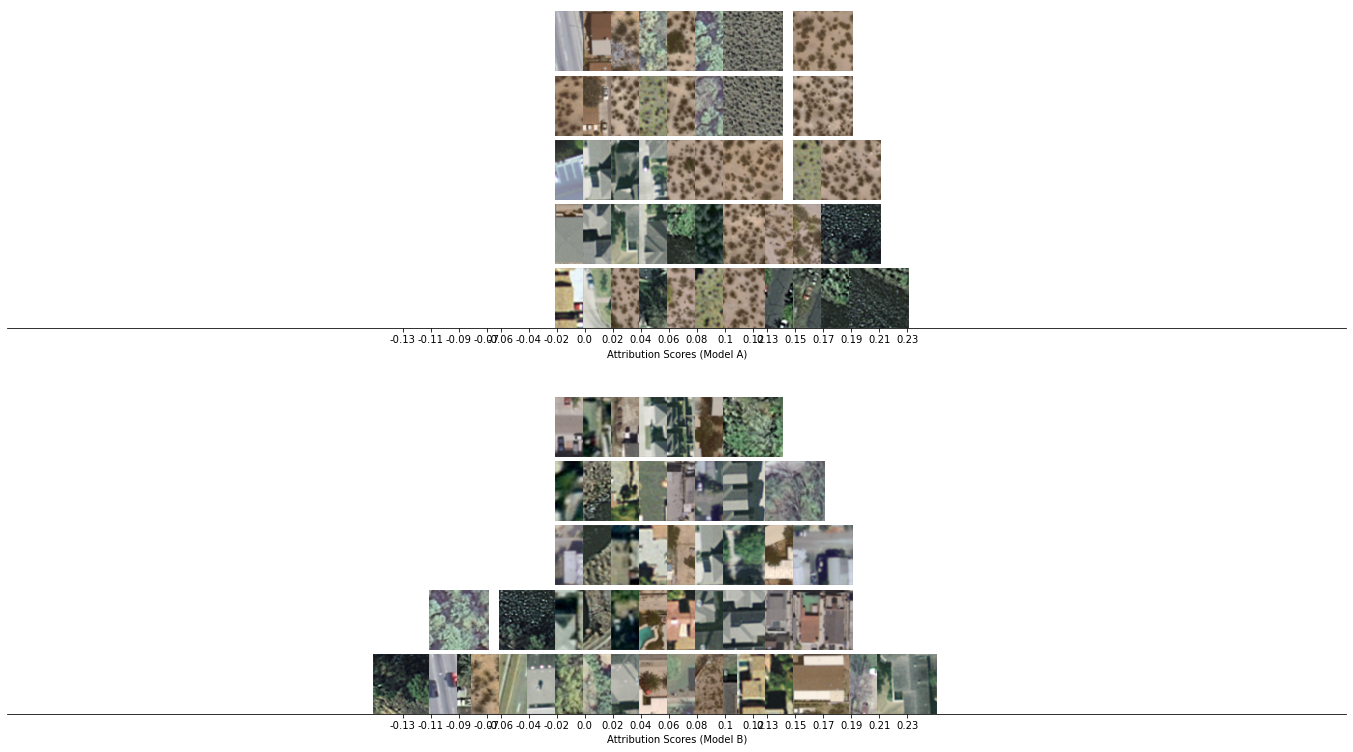}

\end{document}